\documentclass[a4paper, 10pt, conference]{ieeeconf}      % Use this line for a4
\usepackage{FG2026}
\usepackage{amsmath, amssymb, mathrsfs}
\usepackage{cleveref}
\usepackage{graphicx}
\usepackage{amsfonts}

\usepackage{algorithm}
\usepackage{algpseudocode}
\FGfinalcopy % *** Uncomment this line for the final submission
\usepackage[table]{xcolor} % Add this in your preamble
\definecolor{best}{RGB}{135,206,235} % SkyBlue
\definecolor{second}{RGB}{99,184,255}    % LightSkyBlue
\definecolor{third}{RGB}{176,224,230}    % PowderBlue

\IEEEoverridecommandlockouts                              % This command is only
\def\FGPaperID{46} % *** Enter the FG2026 Paper ID here

\title{\LARGE \bf
Combo-Gait: Unified Transformer Framework for Multi-Modal Gait Recognition and Attribute Analysis
}

\author{
	Zhao-Yang Wang$^{1}$
        \;\; Zhimin Shao$^{1}$
        \;\; Anirudh Nanduri$^{2}$
        \;\; Basudha Pal$^{1}$\\
        \;\; Laura McDaniel$^{1}$
	\;\; Jieneng Chen$^{1}$
	\;\; Rama Chellappa$^{1}$ \\
	$^1$Johns Hopkins University
    $^2$
University of Maryland
 \\ 
}

\begin{document}

\ifFGfinal
\thispagestyle{empty}
\pagestyle{empty}
\else
\author{Anonymous FG2026 submission\\ Paper ID \FGPaperID \\}
\pagestyle{plain}
\fi
\maketitle

%%%%%%%%%%%%%%%%%%%%%%%%%%%%%%%%%%%%%%%%%%%%%%%%%%%%%%%%%%%%%%%%%%%%%%%%%%%%%%%%
\begin{abstract}

Gait recognition is an important biometric for human identification at a distance, particularly under low-resolution or unconstrained environments. Current works typically focus on either 2D representations (e.g., silhouettes and skeletons) or 3D representations (e.g., meshes and SMPLs), but relying on a single modality often fails to capture the full geometric and dynamic complexity of human walking patterns. In this paper, we propose a multi-modal and multi-task framework that combines 2D temporal silhouettes with 3D SMPL features for robust gait analysis. Beyond identification, we introduce a multitask learning strategy that jointly performs gait recognition and human attribute estimation, including age, body mass index (BMI), and gender. A unified transformer is employed to effectively fuse multi-modal gait features and better learn attribute-related representations, while preserving discriminative identity cues. Extensive experiments on the large-scale BRIAR datasets, collected under challenging conditions such as long-range distances (up to 1 km) and extreme pitch angles (up to 50°), demonstrate that our approach outperforms state-of-the-art methods in gait recognition and provides accurate human attribute estimation. These results highlight the promise of multi-modal and multitask learning for advancing gait-based human understanding in real-world scenarios.
\end{abstract}

%%%%%%%%%%%%%%%%%%%%%%%%%%%%%%%%%%%%%%%%%%%%%%%%%%%%%%%%%%%%%%%%%%%%%%%%%%%%%%%%
\section{INTRODUCTION}

\label{sec:intro}

\begin{figure}[t]
  \centering
  % \fbox{\rule{0pt}{3in} \rule{1\linewidth}{0pt}}
  {\includegraphics[scale=0.6]{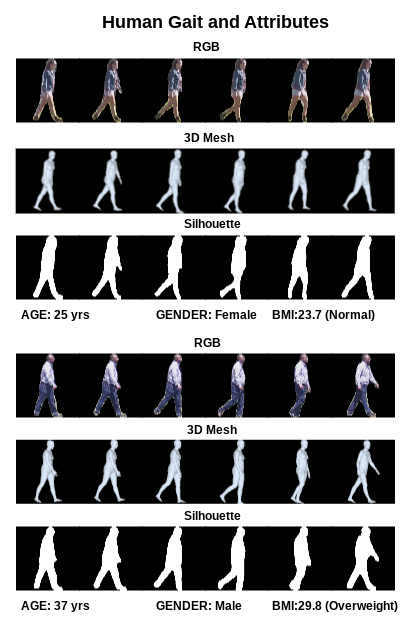}}
   \caption{An example of different gait representations with human attributes information. Human attributes such as Age, BMI and Gender influence a subject's walking pattern and shape. }
   \label{fig:uni_intro}
   \vspace{-4mm}
\end{figure}

Gait recognition stands as a powerful biometric for human identification~\cite{ye2024biggait}, offering a unique advantage in recognizing individuals at a distance without requiring high-resolution imagery or physical contact~\cite{bashir2010gait}. Unlike facial or fingerprint recognition, gait analysis relies on the dynamic patterns of human walking, making it particularly valuable for long-range surveillance and unconstrained environments~\cite{singh2018vision}. However, achieving robust gait recognition under real-world conditions such as low resolution, occlusion, and varying camera viewpoints remains a significant challenge~\cite{wang2024hypergait}. Currently, gait recognition methods have primarily relied on 2D representations, such as silhouettes~\cite{zhang2004human} or skeletons~\cite{teepe2021gaitgraph}, which capture appearance and motion cues from video sequences. While these approaches are computationally efficient and effective in controlled settings, they are sensitive to viewpoint changes and clothing variations~\cite{wang2025unigait}. On the other hand, 3D representations, including body meshes~\cite{li2022multi}, and SMPL parameters~\cite{zheng2022gait}, provide richer structural and geometric information that is inherently viewpoint-invariant. Nevertheless, 3D-only approaches are prone to reconstruction errors and the loss of fine-grained temporal cues from visual appearance, offering only complementary information to silhouette-based representations~\cite{wang2025vm}\cite{zhang2022gait}. Thus, a single representation often fails to fully characterize the complex geometry and dynamics of human gait.

To address these limitations, we propose Combo-Gait, a multi-modal and multi-task gait recognition framework that integrates 2D silhouette features and complementary 3D SMPL-based features to capture both visual and structural aspects of human walking. By combining these modalities, Combo-Gait leverages the strengths of each while mitigating their individual weaknesses, resulting in a more comprehensive and robust gait representation.

Beyond identity recognition, human attribute information provides valuable biometric and demographic information, such as age, gender, and body mass index (BMI). Motivated by this observation, we introduce a multi-task learning strategy that jointly performs gait recognition and human attribute estimation within a unified framework. This joint formulation allows shared representations to benefit both tasks—enhancing discriminative identity features while improving the estimation of physical attributes. Combo-Gait is a unified transformer-based fusion network, which effectively integrates multi-modal gait features and learns attribute-aware representations. The transformer architecture enables global context modeling across temporal sequences and modalities, ensuring that the fused representations retain both discriminative and semantically meaningful information.

We evaluate Combo-Gait on the BRIAR datasets, a large-scale benchmark captured under challenging real-world conditions, including long-range distances (up to 1 km) and extreme pitch angles (up to 50°). Experimental results demonstrate that our proposed framework not only surpasses state-of-the-art methods in gait recognition performance but also achieves accurate estimation of human attributes. These findings highlight the potential of multi-modal and multi-task learning for advancing gait-based human understanding in complex, unconstrained scenarios.

\begin{figure*}[t]
  \centering
  % \fbox{\rule{0pt}{3in} \rule{1\linewidth}{0pt}}
  % \fbox{}
  {\includegraphics[width=1.0\linewidth]{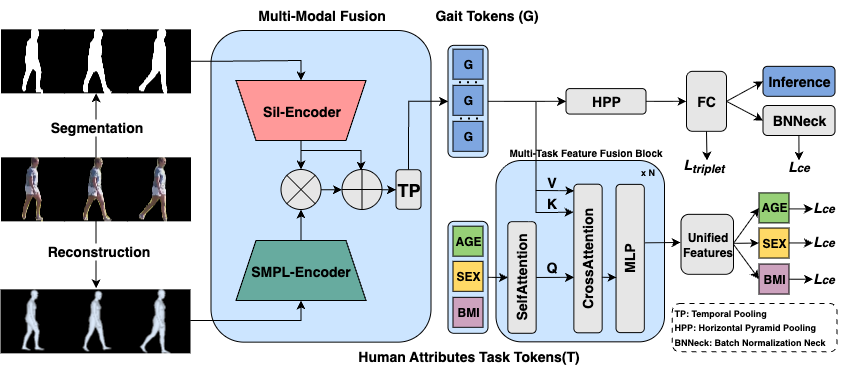}}
   \caption{The pipeline of the Combo-Gait framework. (1) Video Segmentation and Reconstruction; (2) Multimodal Gait Feature Extraction and Fusion; (3) Gait Feature and Human Attribute Fusion; (4) Gait Recognition and Human Attribute Estimation Execution.} 
   \label{fig:Combo-Gait Framework}
   \vspace{-4mm}
\end{figure*}
%%%%%%%%%%%%%%%%%%%%%%%%%%%%%%%%%%%%%%%%%%%%%%%%%%%%%%%%%%%%%%%%%%%%%%%%%%%%%%%%
\section{RELATED WORKS}
\label{sec:formatting}

Gait recognition methods can generally be categorized into appearance-based and model-based approaches~\cite{fan2023opengait}. Appearance-based gait recognition, extracting shape features from video or image sequences, has gained widespread popularity~\cite{fan2020gaitpart}\cite{chao2021gaitset}\cite{wang2003silhouette}\cite{chao2021gaitset}\cite{lin2021gait}\cite{liang2022gaitedge}\cite{fan2023exploring}. Model-based gait recognition methods often use pose~\cite{meng2023fastposegait}\cite{liao2024denseposegait}, skeleton~\cite{teepe2022towards}\cite{fan2024skeletongait}, point cloud~\cite{shen2023lidargait}, virtual marker\cite{wang2025vm}, mesh~\cite{li2022multi} and SMPL~\cite{zheng2022gait} to extract structure features from human body.

Current gait recognition methods have relied more on 2D representations with the boom of deep learning, such as  GaitSet~\cite{chao2021gaitset}, GaitPart~\cite{fan2020gaitpart}, GaitGL~\cite{lin2021gait}, GaitBase~\cite{fan2023opengait} and DeepGaitV2~\cite{fan2023exploring}. While these approaches achieve good recognition performance in controlled settings, they are affected by viewpoint changes, turbulence and noise in real-world applications~\cite{wang2024hypergait}\cite{bertocco2024vision}\cite{myers2025unconstrained}. However, 3D representations, including body meshes~\cite{li2022multi} and SMPL parameters~\cite{zheng2022gait}, can offer richer body structural and geometric information that is inherently viewpoint-invariant. Therefore, integrating complementary information from 3D representations into silhouette-based representations presents a promising avenue for advancement~\cite{shen2023lidargait}\cite{zheng2022gait}\cite{fan2024skeletongait}\cite{wang2025vm}. 

In addition to identity recognition, Develop multi-tasking gait model is getting more and more attention~\cite{wang2024hypergait}\cite{wang2025unigait}\cite{marin2017deep}\cite{delgado2018end}. Enhancing the performance of each subtask can improve the overall system performance~\cite{narayan2024facexformer}\cite{ranjan2017hyperface}\cite{ranjan2017all}. Human attributes, including age~\cite{gabell1984effect}\cite{aderinola2021learning}, gender~\cite{nigg1994gait}, and body mass index (BMI)~\cite{rosso2019influence}, offer valuable biometric insights. Based on the above, in this paper, we propose a multi-modal, multi-task model for performing gait recognition and human attribute estimation. We select silhouettes and 3D SMPL parameters because silhouettes capture detailed visual information, whereas 3D SMPL parameters provide a more compact and efficient representation compared to full 3D meshes.

%%%%%%%%%%%%%%%%%%%%%%%%%%%%%%%%%%%%%%%%%%%%%%%%%%%%%%%%%%%%%%%%%%%%%%%%%%%%%%%%
\section{METHODOLOGY}
\label{sec:methodology}
In this section, we present the details of our proposed Combo-Gait method, as illustrated in ~\Cref{fig:Combo-Gait Framework}. The framework is composed of four main modules: (1) Video Segmentation and Reconstruction; (2) Multimodal Gait Feature Extraction and Fusion; (3) Gait Feature and Gait-Related Human Attribute Fusion; (4) Gait Recognition and  Human Attribute Estimation Execution.

\subsection{Video Segmentation and Reconstruction}
To extract binary silhouette sequences from videos, we apply a segmentation method to obtain silhouette masks: \[\mathbf{X}_{\text{sil}} = \{X_{t}\}^{T}_{t=1} \in \mathbb{R}^{B \times T \times H \times W}\] where $X_{t} \in \mathbb{R} ^{H\times W}$ denotes the silhouette mask at the $t$-th frame, $T$ is the sequence length, $B$ is the batch size, $H$ is the height of each silhouette mask, and $W$ is the width of each silhouette mask.
Similarly, to obtain 3D SMPL parameters, we use a reconstruction method that produces: \[\mathbf{Y}_{\text{smpl}} = \{Y_{t}\}^{T}_{t=1} \in \mathbb{R}^{B,T,D}\] where $D=23 \times 3 + 10 + 1 \times 3 = 82$ denotes the dimension of the SMPL parameter vector, which includes the 3D human body pose ($23 \times 3$), body shape coefficients (10), and global root orientation ($1 \times 3$).

\subsection{Multimodal Gait Feature Extraction and Fusion}
As illustrated in~\Cref{fig:Combo-Gait Framework}, to extract initial gait features from silhouettes, silhouette sequences are fed into a CNN-based encoder to extract silhouette features formulated as \[\mathbf{E}_{\text{sil}} = f_{cnn}(X_{\text{sil}}) \in \mathbb{R}^{B \times C \times T \times H^{'} \times W^{'}}\] where $C$ is the number of feature channels produced by the CNN encoder. In the meantime, a multi-layer perceptron (MLP) projects this vector into a learned SMPL embedding: \[\mathbf{E}_{\text{smpl}} = f_{smpl}(Y_{\text{smpl}}) \in \mathbb{R}^{B \times T \times D^{'}} \] where $D^{'} = 256$ is the embedding dimension, and $E_{smpl}$ denotes three fully connected layers with batch normalization and dropout. 

To fuse extracted silhouette feature and smpl feature, we reshape  $\mathbf{E}_{\text{smpl}}$ into a matrix: $\mathbf{E}_{\text{smpl}}\in \mathbb{R}^{B \times 1 \times T \times H^{'} \times W^{'}}$. Then broadcast the SMPL feature along the channel dimension so that it matches the silhouette feature tensor’s channel size $C$. \[\mathbf{E}_{\text{smpl}} = repeat(\mathbf{E}_{\text{smpl}}, C)\in \mathbb{R}^{B \times C \times T \times H^{'} \times W^{'}}\]

For computational convenience, we expand $\mathbf{E}_{\text{sil}}$ to square matrices by zero-padding along the shorter edge: 
\[\mathbf{E}_{\text{sil}} \xrightarrow{\text{ZeroPad}} \mathbf{E}_{\text{sil}}^{'},  \in \mathbb{R}^{B \times C \times T \times H^{'} \times H^{'}}\] where $H^{'} = \max(H^{'},W^{'})$.
The fusion is then performed via:
\[
\mathbf{E}_{\text{fused}} 
= \mathbf{E}_{\text{sil}}' 
\cdot 
\left( \mathbf{I} + 
 \mathbf{E}_{\text{smpl}}
\right)
\]
where $\mathbf{I} \in \mathbb{R}^{H' \times H'}$, is the identity matrix, and 
$\mathbf{E}_{\text{fused}} 
\in \mathbb{R}^{B \times C' \times T \times H' \times H'}$ denotes the fused feature representation.

Temporal Pooling (TP) is used to aggregate time frame-level features: \[{\mathbf {G}_{\text{fused}}} = f_{TP}(\mathbf{E}_{\text{fused}}) \in \mathbb{R}^{B,C',H',H'}\]

\subsection{Gait Feature and Human Attribute Fusion}
To achieve effective multi-task learning across different domains, we propose a unified transformer architecture for human attributes estimation. Each attribute prediction is associated with a distinct \emph{task token} $\mathbf{T}$, while the fused gait feature is represented by a set of \emph{gait tokens} $\mathbf{G}$.

Let $\mathbf{T}$ denote the set of attribute tasks (e.g., Age, BMI, Gender). For each task \( t_k \), we define a learnable token embedding: $\mathbf{t}_k \in \mathbb{R}^M, \quad k = 1, 2, \dots J$ , where $M$ is the transformer embedding dimension and \(J\) is the number of tasks.
All task tokens are concatenated into a task token matrix:
$
\mathbf{T} = 
[\mathbf{t}_{\text{age}},
\mathbf{t}_{\text{sex}},
\mathbf{t}_{\text{bmi}}]
\in \mathbb{R}^{B \times J \times M}.
$
These task tokens are initialized randomly and learned during training:
\[\mathbf{t}_k \sim \mathcal{N}(0, \sigma^2)\] 
The gait tokens are derived from the fused gait features ${\mathbf {G}_{\text{fused}}}$. To fuse gait feature and human attribute feature, we design a Multi-Task Feature Fusion Block. In each block, the attribute task tokens are first passed through a self-attention module to enhance the task-specific representations and improve inter-task synergy. $\mathbf{T}$ is updated as: 
\[{\bf T^{'}} = \mathrm{SelfAttn}(Q={\bf T^{'}}, K = {\bf T, V={\bf T )}}\in \mathbb{R}^{B\times J \times M} \] 

where $Q$, $K$, and $V$ represent the queries, keys, and values, respectively. To allow each attribute representation $\mathbf{T}'$ to interact with the fused gait features $\mathbf {G}_{\text{fused}}$, we apply a cross attention module: 
\[{\bf \hat{T}} = \mathrm{CrossAttn}(Q={\mathbf{T'}, K = {\mathbf {G}_{\text{fused}}}, V={\mathbf {G}_{\text{fused}}}})\] where ${\bf \hat{T}}\in \mathbb{R}^{B\times J \times M}$.
A multi-layer perceptron (MLP) with two linear layers is then employed to obtain unified prediction heads for each attribute: \[\tilde{\mathbf{T}} = f_{mlp}(\mathbf {\hat{T}})\in \mathbb{R}^{B\times J \times M}\] Meanwhile, the gait tokens are processed through a Horizontal Pyramid Pooling (HPP) module to extract discriminative gait features: \[\tilde{\mathbf{G}} = f_{\text{HPP}}(\mathbf{G}_{\text{fused}}) \in \mathbb{R}^{B \times C' \times H'}\]

\subsection{Gait Recognition and Human Attribute Estimation Execution}
For gait recognition, the refined gait features $\tilde{\mathbf{G}}$ are fed into a fully connected layer to produce the final gait representation:  \[\mathbf {F_{gait}} = f_{fc}(\tilde{\mathbf{G}})\in \mathbb{R}^{B\times C'' \times P}\] where $P$ represents the number of body parts. 

For human attribute estimation, the refined task-specific representaions are denoted as: $\tilde{\mathbf{T}} = [\tilde{\mathbf{T}}_{age},\tilde{\mathbf{T}}_{sex},\tilde{\mathbf{T}}_{bmi}]$. Each attribute token is passed through a corresponding fully connected layer to produce task-specific outputs:

$\mathbf {F_{age}} = f_{fc}(\tilde{\mathbf{T}}_{age})\in \mathbb{R}^{B\times Q_{age}}; \quad
\mathbf {F_{sex}} = f_{fc}(\tilde{\mathbf{T}}_{sex})\in \mathbb{R}^{B\times Q_{sex}}; \quad
\mathbf {F_{bmi}} = f_{fc}(\tilde{\mathbf{T}}_{bmi})\in \mathbb{R}^{B\times Q_{bmi}}$
where $Q_{age}$,$Q_{sex}$,$Q_{bmi}$ represent the number of classes for age, sex, and bmi estimation, respectively.

\subsection{Loss Function}
The overall objective combines cross-entropy and triplet losses for gait recognition and human attribute estimation. The total loss is formulated as:

\[
\mathscr{L}_{\text{Combo-Gait}}
= \alpha (\mathscr{L}_{\text{tri}}^{\text{gait}} + \mathscr{L}_{\text{ce}}^{\text{gait}})
+ \beta \mathscr{L}_{\text{ce}}^{\text{task}},
\]
and can be expanded as:
\[
\mathscr{L}_{\text{Combo-Gait}}
= \alpha_1 \mathscr{L}_{\text{tri}}^{\text{gait}}
+ \alpha_2 \mathscr{L}_{\text{ce}}^{\text{gait}}
+ \beta_1 \mathscr{L}_{\text{ce}}^{\text{age}}
+ \beta_2 \mathscr{L}_{\text{ce}}^{\text{sex}}
+ \beta_3 \mathscr{L}_{\text{ce}}^{\text{bmi}}
\]

Each component is scaled by corresponding weights $\alpha_i$ and $\beta_i$ to balance the relative influence of gait recognition and human attribute estimation during joint training.

\section{EXPERIMENTS}
\label{sec:experiments}
\subsection{Datasets}
\begin{figure}[t]
  \centering
  % \fbox{\rule{0pt}{3in} \rule{1\linewidth}{0pt}}
  % \fbox{}
  {\includegraphics[width=1.0\linewidth]{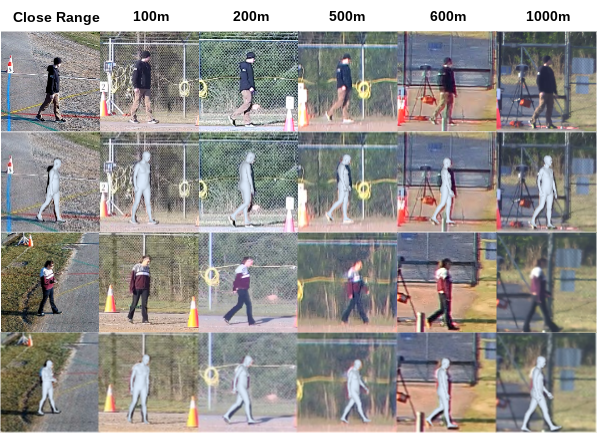}}
   \caption{Examples of two subjects under various conditions from the BRIAR dataset~\cite{briar}. At longer distances, significant turbulence and noise degrade the image quality. The 3D SMPL parameters corresponding to the two subjects are presented in the second and fourth rows. }
   \label{fig:briardataset}
   \vspace{-7mm}
\end{figure}

The BRIAR Dataset~\cite{cornett2023expanding}:
In our paper, we evaluate the proposed Combo-Gait model using three large-scale datasets from the Biometric Recognition and Identification at Altitude and Range (BRIAR) program~\cite{bolme2024data, cornett2023expanding}: BGC1, BGC2, and BGC3. The BRIAR datasets are designed to facilitate research on human biometric recognition under highly challenging conditions, including varying ranges (from close range up to 1,000 meters) and steep pitch angles of up to $50^{\circ}$. Representative examples of subjects captured under different conditions from the BRIAR datasets are shown in~\Cref{fig:briardataset}. Each dataset, BGC1, BGC2, and BGC3, contains multiple modalities, including close-range and field long-range silhouettes (captured at 100 meters, 200 meters, …, 1,000 meters), 3D SMPL parameters, and human attribute meta information (age, gender and bmi).

The training set includes data from BGC1 (158 subjects), BGC2 (194 subjects), and BGC3 (170 subjects), for a total of 522 subjects and 34,501 videos. The test set includes BGC1 (64 subjects), BGC2 (98 subjects), and BGC3 (82 subjects), for a total of 244 subjects and 3,196 videos. These datasets provide a comprehensive evaluation of the proposed Combo Gait network in real-world environments that involve atmospheric turbulence, noisy imaging, occlusions, long-range capture, clothing variations, and multiple camera viewpoints.

Moreover, the gait-related human attribute metadata in the BGC1–BGC3 datasets includes: Age: 18-85 years;
Height: 52-81 inches.
Weight: 93-438 lbs and BMI: 14.23-68.65.
BMI Status: Underweight, Healthy, Overweight, and Obese.
Gender: Classified into two categories: female and male.

\begin{table*}[t]
\small
\centering
\renewcommand{\arraystretch}{1.25}
\setlength{\tabcolsep}{3pt}
\scalebox{1.0}{
\begin{tabular}{c|cccccccccc}
\hline
\textbf{Methods} & Rank1 & Rank2 & Rank3 & Rank4 & Rank5 & Rank6 & Rank7 & Rank8 & Rank9 & Rank10 \\ \hline

GaitSet & 36.68 & 46.24 & 53.11 & 58.40 & 63.60 & 66.02 & 70.01 & 72.14 & 73.44 & 74.93 \\
GaitGL & 40.58 & 51.90 & 58.96 & 63.70 & 67.13 & 71.03 & 72.98 & 75.02 & 76.51 & 77.53 \\
GaitPart & 42.25 & 51.62 & 58.12 & 63.51 & 67.87 & 70.47 & 72.24 & 74.28 & 76.42 & 77.62 \\
UniGait & 44.48 & 54.13 & 60.35 & 64.90 & 67.50 & 70.57 & 73.17 & 75.21 & 76.51 & 77.81 \\
GaitBase & 51.90 & 63.05 & 68.34 & 73.35 & 76.14 & 79.02 & 81.99 & 83.10 & 84.59 & 85.98 \\
SMPLGait & 54.22 & 66.95 & 74.09 & \cellcolor{third}78.46 & \cellcolor{third}81.89 & 83.19 & \cellcolor{third}85.33 &\cellcolor{third} 87.47 & \cellcolor{third}88.49 & \cellcolor{third}90.06 \\
SwinGait & 57.75 & \cellcolor{third}68.62 & \cellcolor{third}74.28 & 77.25 & 79.67 & 82.54 & 83.94 & 85.24 & 86.35 & 88.39 \\
DeepGaitV2 & \cellcolor{third}58.96 & 67.50 & 73.82 & 77.25 & 80.97 & \cellcolor{third}83.66 & 84.68 & 86.07 & 87.00 & 88.58 \\
\textbf{Combo-Gait} & \cellcolor{best}\textbf{68.06} & \cellcolor{best}\textbf{77.25} & \cellcolor{best}\textbf{82.82} & \cellcolor{best}\textbf{85.98} & \cellcolor{best}\textbf{87.65} & \cellcolor{best}\textbf{89.69} & \cellcolor{best}\textbf{90.25} & \cellcolor{best}\textbf{91.18} & \cellcolor{best}\textbf{92.11} & \cellcolor{best}\textbf{92.76} \\
\hline
\end{tabular}}

\vspace{-2mm}
\end{table*}

\begin{table*}[]
\small
 \centering
     \renewcommand{\arraystretch}{1.3}
\begin{tabular}{c|cccc}
\hline
Methods & Accu\_Age\_Class & Accu\_BMI\_Class & Accu\_Sex\_Class  \\ \hline
UniGait  & \cellcolor{third}34.42 & \cellcolor{third}64.83 & \cellcolor{third}85.61\\
\textbf{Combo-Gait}  & \cellcolor{best}\textbf{52.25}	& \cellcolor{best}\textbf{72.84}  &\cellcolor{best} \textbf{91.11}               \\ \hline
\end{tabular}

\caption{Gait Recognition (\%) and Human Attribute Estimation (\%) Results on the BRIAR dataset. We compare our proposed Combo-Gait with eight state-of-the-art methods. Accu\_Age\_Class represents the classification accuracy for the predicted age group. For brevity, Age is used to denote Accu\_Age\_Class in all subsequent tables. The best and second-best results are highlighted in dark and light blue.}

\label{human}
\vspace{-2mm}
\end{table*}

\begin{table*}[]
\small
 \centering
   \renewcommand{\arraystretch}{1.3}
\scalebox{1.0}
{\begin{tabular}{c|cccccccccc}
\hline
Methods  & Rank1  & Rank2  & Rank3  & Rank4  & Rank5  & Rank6  & Rank7  & Rank8  & Rank9  & Rank10       \\ \hline

Close Range & 75.20 & 80.42 & 83.81 & 87.47 & 88.77 & 89.82 & 90.34 & 91.12 & 92.43 & 93.21 
\\
100m & 60.58 & 75.96 & 78.85 & 81.73 & 84.62 & 87.50 & 88.46 & 90.38 & 92.31 & 93.27 \\
200m & 63.16 & 77.19 & 80.70 & 80.70 & 80.70 & 84.21 & 84.21 & 87.72 & 87.72 & 87.72 \\
400m & 62.02 & 72.09 & 85.27 & 86.05 & 87.60 & 87.60 & 87.60 & 87.60 & 89.15 & 89.15 \\
500m & 64.71 & 74.12 & 80.00 & 82.35 & 83.53 & 90.59 & 92.94 & 92.94 & 92.94 & 92.94 \\
600m & 54.64 & 69.07 & 76.29 & 83.51 & 85.57 & 87.63 & 88.66 & 89.69 & 90.72 & 90.72 \\
1000m & 48.15 & 55.56 & 74.07 & 81.48 & 81.48 & 85.19 & 85.19 & 85.19 & 85.19 & 92.59 
\\

\hline

\end{tabular}
}
\caption{Gait Recognition (\%) Under Varying Distances. Combo-Gait maintains robust performance even at extreme ranges.}

\label{distance_results}
 \vspace{-5mm}
\end{table*}

\subsection{Comob-Gait Implementation Details}

Video Segmentation and 3D Reconstruction:
We use the silhouette sequences and 3D SMPL parameters provided from the BRIAR datasets, The input binary silhouettes are preprocessed to a fixed resolution of 
$H=64$ and $W = 44$ While the number of frames varies across silhouette sequences, training uses a fixed selection of $30$ frames per silhouette sequence. The 3D SMPL parameters has a total dimension of $82$ corresponding each binary silhouette, consisting of the 3D human body pose ($23 \times 3$), body shape coefficients (10), and global root orientation ($1 \times 3$). During training, the batch size is set to $16 \times 4$, where $16$ the number of selected subjects and $4$ represents the number of silhouette sequences sampled per subject. The model is trained on 8 NVIDIA A5000 GPUs, resulting in an effective batch size of $8$ per GPU. The input silhouettes and corresponding 3D SMPL parameters are represented as: $\mathbf{X}_{\text{sil}} \in \mathbb{R}^{B \times T \times H \times W} = \mathbb{R}^{8 \times 30 \times 64 \times 44}$ $\mathbf{Y}_{\text{smpl}} \in \mathbb{R}^{B,T,D}= \mathbb{R}^{8 \times 30 \times 82}$

Multimodal Gait Feature Extraction and Fusion:
For the Sil-Encoder $f_{cnn}(\cdot)$, we use DeepGaitV2~\cite{fan2023exploring} as the backbone for the Combo-Gait framework. DeepGaitV2, a state-of-the-art gait recognition encoder, effectively extracts robust gait features from the silhouette sequences. The output feature $
\mathbf{E}_{\text{sil}}\in \mathbb{R}^{B \times C \times T \times H' \times W'} = \mathbb{R}^{8 \times 512 \times 30 \times 16 \times 11}$
 To demonstrate the flexibility of our framework, we also experiment with different $cnn$-based encoders used in gait recognition, providing a comparison of performance across various feature extraction methods. For the SMPL-Encoder $f_{smpl}(\cdot)$ consist of three fully connected layers with batch normalization and dropout. The output feature $\mathbf{E}_{\text{smpl}} \in \mathbb{R}^{B \times T \times D^{'}} = \mathbb{R}^{8 \times 30 \times 256}$

Gait Feature and Human Attribute Fusion:
To enable both self-attention and cross-attention between gait tokens and human attribute task tokens, the multi-task feature fusion block integrates key modules partly adapted from the FaceXformer~\cite{narayan2024facexformer} architecture. The number of block is set to 2. The output feature from the $ \mathrm{SelfAttn}$ module is ${\bf T^{'}} \in \mathbb{R}^{B\times J \times M} = \mathbb{R}^{8\times 3 \times 512}$ and the output feature from the $ \mathrm{CrossAttn}$ module is ${\bf \hat{T}} \in \mathbb{R}^{B\times J \times M} = \mathbb{R}^{8\times 3 \times 512}$. The unified prediction head, implemented through the $f_{mlp}(\cdot)$ function, produces $\tilde{\mathbf{T}} \in \mathbb{R}^{B\times J \times M} = \mathbb{R}^{8\times 3 \times 512}$. The gait tokens are processed through $f_{\text{HPP}(\cdot)}$ module, resulting in $\tilde{\mathbf{G}} \in \mathbb{R}^{B \times C' \times H'} = \mathbb{R}^{8 \times 512 \times 16}$.
Gait Recognition and Human Attribute Estimation Execution: The final gait representation and human attribute representations after separate $f_{fc}(\cdot)$ functions are: $\mathbf {F_{gait}} \in \mathbb{R}^{B\times C'' \times P}= \mathbb{R}^{8\times 256 \times 16}$; For human attribute estimation, age is divided into five 20-year intervals; BMI is categorized into four groups: underweight, healthy, overweight, and obese; and sex is classified into two categories: female or male.The corresponding final representations are: $\mathbf {F_{age}}  \in \mathbb{R}^{B\times Q_{age}} = \mathbb{R}^{8\times 5}; 
\mathbf {F_{sex}}  \in \mathbb{R}^{B\times Q_{sex}} = \mathbb{R}^{8\times 2}; 
\mathbf {F_{bmi}}  \in \mathbb{R}^{B\times Q_{bmi}} = \mathbb{R}^{8\times 4}$.

\subsection{Training and Testing Strategy}

For comprehensive evaluation and comparison with other gait recognition methods, we trained and test our experiments based on the OpenGait~\cite{fan2023opengait} platform. Combo-Gait is training in total 200,000 iterations using stochastic gradient descent (SGD) as the optimizer, with a learning rate of 0.01, momentum of 
0.9, and weight decay of $5 \times 10^{-4}$. 

To assess the proposed model on the BRIAR dataset, we merge the three constituent datasets (BGC1, BGC2, and BGC3) for both training and evaluation. All samples are gathered together to form a single training set, and likewise combined to create a testing set. Evaluation adopts a commonly used Probe-Gallery paradigm: the gallery contains labeled reference samples, while the probe set comprises queries whose identities must be matched against the gallery.

Gait recognition is quantified using Rank-N accuracy (Rank-1 to Rank-10), indicating the proportion of correct matches within the top N ranked predictions. This evaluation approach aligns with prior gait recognition studies such as SMPLGait~\cite{zheng2022gait}, UniGait~\cite{wang2025unigait}, and OpenGait~\cite{fan2023opengait}.

For human attribute estimation, performance is reported as the mean classification accuracy across attributes. Specifically, we evaluate the accuracy of predicting age, gender and bmi. This dual evaluation framework offers a holistic measure of the model’s capabilities, covering both gait recognition and attribute estimation.

\begin{table*}[]
 \centering
 \renewcommand{\arraystretch}{1.3}
\scalebox{1.0}
{\begin{tabular}{c|cccccc}
\hline
Methods  & Rank1  & Rank5    & Rank10   & Age & BMI & Sex     \\ \hline
SMPLGait & 54.69 & 78.18  & 86.44&-&-&-\\
\textbf{Combo-Gait w SMPLGait} & \textbf{56.92} & \textbf{81.24}  & \textbf{89.04}&48.53&68.30&89.58 \\\hline
GaitBase & 51.90 & 76.14  & 85.98&-&-&- \\
\textbf{Combo-Gait w GaitBase} & \textbf{58.12} & \textbf{82.36}  & \textbf{90.06}	&48.69&69.15&89.58   \\ \hline
DeepGaitV2 & 58.96  & 80.97  & 88.58&-&-&-\\
\textbf{Combo-Gait w DeepGaitV2} & \textbf{68.06} & \textbf{87.65} & \textbf{92.76}&\textbf{52.25}&\textbf{72.84}&\textbf{91.11}\\

\hline

\end{tabular}
}

\caption{Gait Recognition (\%) and Human Attribute Estimation (\%) with three state-of-the-art silhouette backbones. The Combo-Gait is flexible and effective with different backbones. }
\label{human_backbone}
\vspace{-8mm}
\end{table*}

\section{Experimental Results}
\label{sec:results}
\subsection{Comparison with Other State-of-the-Arts}
We evaluate the performance of the proposed Combo-Gait framework for gait recognition and human attribute estimation, comparing it against eight state-of-the-art single-modality and multi-modality methods. As shown in Table~\ref{human}, Combo-Gait consistently achieves superior performance across all ranks, attaining a Rank-1 accuracy of 68.06\%, which significantly surpasses the single-modality method (DeepGaitV2~\cite{fan2023exploring}) and the multi-modality(SMPLGait~\cite{zheng2022gait}). Moreover, Combo-Gait achieves the highest accuracy across all ranking levels up to Rank-10, reaching 92.76\% at Rank-10. These results demonstrate the effectiveness of our approach in integrating multi-modality gait features, allowing the 3D SMPL parameters to provide complementary information to silhouette-based representations. This integration significantly enhances robustness and accuracy, enabling reliable gait recognition in highly challenging real-world conditions such as turbulence, long distances, occlusions, and varied viewing angles. Table~\ref{human} also reports classification accuracy for three human attributes: Age, BMI, and Sex. Combo-Gait significantly outperforms UniGait across all attribute categories, achieving 52.25\% in age classification, 72.84\% in BMI classification, and 91.11\% accuracy in sex classification. This confirms that our model effectively captures rich attribute-related information, enhancing the robustness and applicability of the framework for attribute estimation alongside gait recognition. This demonstrates a clear multi-tasking advantage, where simultaneous gait recognition and attribute estimation mutually reinforce each other, improving overall system performance. Together, these results illustrate the multi-modal benefit of leveraging complementary data sources (silhouettes and 3D SMPL parameters) and the multi-task benefit of jointly learning gait recognition and human attribute estimation. The synergy between these two aspects enables Combo-Gait to deliver superior performance, robustness, and generalizability, making it a strong candidate for real-world gait analysis and biometric applications.

\subsection{Performance Under Varying Distances}
Table~\ref{distance_results} presents gait recognition performance across different distances in the BRIAR datasets. Performance generally decreases as distance increases, highlighting the challenges of long-range gait recognition. Combo-Gait maintains robust performance even at extended ranges. For close-range sequences, Rank-1 accuracy reaches 75.20\%, while at 1000 meters, it still achieves 48.15\% for Rank-1 and 92.59\% for Rank-10. Notably, performance degradation is more gradual at mid-range distances (100m–600m), demonstrating the model’s resilience to distance variation and environmental challenges such as turbulence, noise, and occlusion.

\begin{table}[]
 \centering
 \renewcommand{\arraystretch}{1.3}
\scalebox{1.0}
{\begin{tabular}{c|cccccc}
\hline
Loss Weight   & Rank1  & Rank5  & Rank10  &  Age   &  BMI  &  Sex         \\ \hline

$\alpha$:1.0 $\beta$:0.0 & 63.51 & 83.57 & 90.34 &-&-&-\\
$\alpha$:1.0 $\beta$:0.01 & 68.06 & \textbf{87.65} & \textbf{92.76}& \textbf{52.25}& \textbf{72.84}&91.11 \\
\textbf{$\alpha$:1.0 $\beta$:0.1} & \textbf{68.99} & 86.44 & 91.92&49.69&65.64& \textbf{91.36}\\
$\alpha$:1.0 $\beta$:1.0 & 63.70 & 84.59 & 91.55 &48.65&66.05&90.58\\
\hline

\end{tabular}
}

\caption{Gait Recognition (\%) and Human Attribute Estimation (\%) with different loss weighting}
\label{human_prop}
\vspace{-8mm}
\end{table}

\section{Ablation Study}
To better understand the contributions of key components in the proposed Combo-Gait framework, we conduct a series of ablation studies. We systematically evaluate: The choice of backbone architecture, The effect of multi-task loss weighting,The role of multi-modal feature fusion,The impact of attention mechanisms.Architectural hyperparameters including depth, number of attention heads, and fully-connected (FC) dimensions.

\subsection{Advantages of Multi-Modalities Fusion}
To investigate the advantages of integrating multiple modalities, we examine the performance of Combo-Gait when 3D SMPL parameters combining different silhouette backbone encoders. Table~\ref{human_backbone} shows that incorporating silhouette-based backbones with the 3D SMPL parameters consistently improves performance for all tested architectures (SMPLGait, GaitBase, DeepGaitV2). This confirms that multi-modalities fusion effectively leverages complementary information. Silhouette features capture shape and texture cues, while 3D SMPL parameters encode dynamic pose and shape information that is invariant to viewpoint and appearance, providing complementary information that silhouettes alone cannot capture.

Importantly, the improvements across all backbones demonstrate that the Combo-Gait framework is flexible and backbone-agnostic: it can seamlessly integrate with different silhouette-based architectures to enhance gait recognition performance. Among the tested configurations, the combination of DeepGaitV2 with our fusion mechanism achieves the highest Rank-1 accuracy of 68.06\%, a substantial improvement over its baseline. This validates that multi-modal fusion not only enhances recognition accuracy but also strengthens the adaptability of the framework to various backbone designs.

Table~\ref{human_backbone} further supports this finding, showing consistent gains in human attribute estimation accuracy across all backbones. These results highlight the power of combining different modalities, establishing multi-modal fusion as a key driver of performance in both gait recognition and human attribute estimation.

\subsection{Multi-Task Loss Weighting}
In our framework, we use a multi-task learning loss to jointly optimize gait recognition and human attribute estimation. In this study, we fixed $\alpha = 1.0$ and varied $\beta$ to evaluate how different attribute weighting affects performance (Table~\ref{human_prop}). Increasing $\beta$ generally improves gait recognition performance up to an optimal value, after which performance drops. Setting $\beta = 0.01$ achieves the best trade-off between gait recognition and attribute estimation, yielding the highest Rank-10 accuracy of 92.76\%. Table~\ref{human_prop} shows that $\beta = 0.01$ also delivers superior attribute classification performance, particularly for age (52.25\%) and BMI (72.84\%). This highlights the benefit of balanced multi-task training.

\begin{table*}[]
\small
 \centering
   \renewcommand{\arraystretch}{1.3}
\scalebox{1.0}
{\begin{tabular}{c|cccccc}
\hline
Feature Fusion   & Rank1   & Rank5   & Rank10 & Age  &  BMI   &Sex       \\ \hline	

$\times$ &64.81 & 84.68  & 91.83 &46.37&63.42&90.55	\\
$\checkmark$ & \textbf{68.06} & \textbf{87.65} & \textbf{92.76}&\textbf{52.25}&\textbf{72.84}&\textbf{91.11}\\
\hline

\end{tabular}
}

\caption{Gait Recognition (\%) and Human Attribute Estimation (\%) results w/ or w/o feature fusion}
\label{human_FFwo}
\vspace{-6mm}
\end{table*}

\begin{table*}[]
\small
 \centering
   \renewcommand{\arraystretch}{1.3}
\scalebox{1.0}
{\begin{tabular}{cc|cccccc}
\hline
SelfAttention&CrossAttention   & Rank1   & Rank5  & Rank10 & Age & BMI & Sex       \\ \hline

$\times$&$\checkmark$&66.39 & 87.19  & 92.11&50.78&\textbf{72.93}&91.02	\\
$\checkmark$&$\checkmark$& \textbf{68.06} & \textbf{87.65} & \textbf{92.76}&\textbf{52.25}&72.84&\textbf{91.11}\\

\hline

\end{tabular}

}
\caption{Gait Recognition (\%) and Human Attribute Estimation (\%) results of attention mechanisms in the multi-task fusion}

\label{Human_SACA}
\vspace{-8mm}
\end{table*}

\subsection{The Effectiveness of Multi-Task Feature Fusion}
To evaluate the effectiveness of feature fusion for gait recognition and human attribute estimation within our multi-task learning framework, we conduct two comparative experiments that analyze how fusing gait features with task-specific representations impacts overall performance.

In the baseline setting, we remove both the $\mathrm{SelfAttn}$ and $\mathrm{CrossAttn}$ operations used for fusing gait tokens and attribute task tokens. Instead, we directly use the gait feature representation $\tilde{\mathbf{G}}$ for both gait recognition and human attribute estimation. Specifically, gait recognition is performed through a single fully connected layer, while three separate MLPs are used for predicting age, BMI, and sex. Each MLP consists of three fully connected layers, each followed by batch normalization and ReLU activation. In the fused setting, we retain the original configuration shown in \Cref{fig:Combo-Gait Framework}, where both $\mathrm{SelfAttn}$ and $\mathrm{CrossAttn}$ modules are applied to jointly learn and integrate multi-task representations.

The recognition results are summarized in \Cref{human_FFwo}, demonstrating that the proposed fusion mechanism facilitates effective information exchange between the gait and attribute branches, allowing the network to capture deeper correlations between gait dynamics and human physical characteristics, thereby improving both overall gait recognition accuracy and human attribute estimation performance.

\subsection{The Effectiveness of Attention Mechanisms for Multi-Task Fusion}
As we have demonstrated the benefits of multi-task feature fusion, we further investigate the contribution of the attention mechanisms used in our Combo-Gait framework. Specifically, we analyze the impact of the Self-Attention and Cross-Attention modules on gait recognition and human attribute estimation. The Self-Attention module is designed to enhance intra-modal feature interactions by refining the attribute tokens within each modality. In contrast, the Cross-Attention module facilitates inter-modal information exchange, allowing gait features to be dynamically modulated by human attribute cues and vice versa.

As shown in Table~\ref{Human_SACA}, incorporating Cross-Attention alone enhances both gait recognition and human attribute estimation compared to the model without feature fusion in Table~\ref{human_FFwo}, improving Rank-1 accuracy from 64.81\% to 66.39\%. When both Self-Attention and Cross-Attention are jointly applied, the performance further increases to 68.06\% in Rank-1 accuracy and 92.76\% in Rank-10 accuracy. Similarly, attribute estimation benefits from the complete attention configuration, especially age and gender classification.

These findings demonstrate that attention mechanisms play a critical role in effectively integrating multi-modal and multi-task representations. The Self-Attention module enhances intra-modal feature consistency, while the Cross-Attention module strengthens inter-modal information exchange, enabling the network to learn more discriminative and generalizable representations for both gait recognition and attribute estimation.

\begin{table}
\small
 \centering
   \renewcommand{\arraystretch}{1.3}
\scalebox{1.0}
{\begin{tabular}{c|cccccc}
\hline
\#Block   & Rank1 & Rank5 & Rank10&Age & BMI & Sex        \\ \hline	

1 &66.76 & 86.07  & 92.39&50.31&71.78&91.30\\
\textbf{2} &  \textbf{68.06} & \textbf{87.65} & \textbf{92.76}&\textbf{52.25}&\textbf{72.84}&91.11\\
3 &66.85 & 86.54 & 92.39&51.66&72.53&\textbf{91.61}\\
\hline
\#Head   & Rank1  & Rank5    & Rank10 & Age&BMI & Sex       \\ \hline

2	&65.27 & 86.17  & 92.39&\textbf{52.25}&72.34&\textbf{91.11}\\
\textbf{4} &\textbf{68.06} & \textbf{87.65} & \textbf{92.76}&\textbf{52.25}&\textbf{72.84}&\textbf{91.11}\\	
8 &66.30 & 87.56  & 92.11&48.37&70.93&90.68\\
\hline
\#Dim   & Rank1    & Rank5    & Rank10 &Age&BMI&Sex      \\ \hline	
256 &66.39 & 86.72  & 92.48&51.72&72.59&90.77\\
\textbf{512} & \textbf{68.06} & 87.65  & \textbf{92.76}&\textbf{52.25}&\textbf{72.84}&\textbf{91.11}\\
1024 &67.41  & \textbf{88.21} & 92.57&50.53&71.37&90.33\\\hline

\end{tabular}
}
\caption{The influence of Hyperparameters}

\label{human_depth}
\vspace{-11mm}
\end{table}

\begin{figure*}
  \centering
  % \fbox{\rule{0pt}{3in} \rule{1\linewidth}{0pt}}
  % \fbox{}
  {\includegraphics[width=0.8\linewidth]{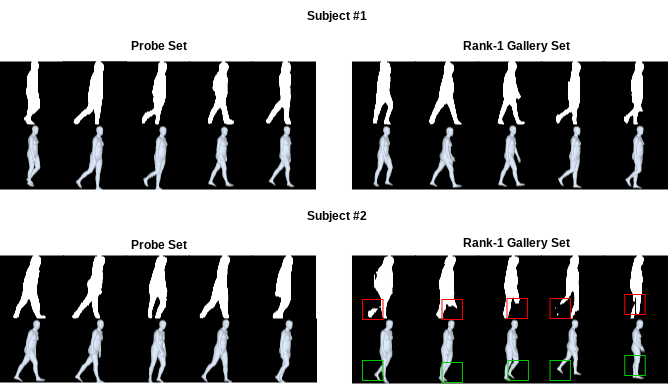}}
   \caption{Visualization of Complementarity between Silhouettes and 3D SMPL parameters}
   \vspace{-5mm}
   \label{fig:Human Attributes Estimation Visualization}
   
\end{figure*}

\subsection{Analysis of Hyperparameters in the Multi-Task Feature Fusion Blocks}
To further examine how hyperparameter choices affect the performance of Combo-Gait, we conducted a systematic ablation study. We examine three key parameters of the multi-task feature fusion block: the number of fusion blocks (\#Block), the number of attention heads (\#Head), and the fully connected layer dimension(\#Dim). The quantitative results are summarized in Table~\ref{human_depth}.

{\bf Effect of multi-task feature fusion block number(\#Block):}
As shown in~\Cref{fig:Combo-Gait Framework}, each multi-task feature fusion block contains a self-attention, a cross-
attention, and an MLP. Increasing the block number from 1 to 2 leads to a notable improvement in gait recognition, achieving the highest Rank-1 accuracy of 68.06\% and Rank-10 accuracy of 92.76\%, as shown in Table~\ref{human_depth}. This configuration also delivers the best performance in human attribute estimation for age (52.25\%) and BMI (72.84\%) prediction. However, further increasing the depth to 3 does not bring additional gains and slightly degrades gait recognition performance, likely due to model overfitting or redundancy in representation learning. Therefore, a depth of 2 offers the best trade-off between complexity and performance.

{\bf Effect of Attention Heads (\#Head):}
We further analyze the influence of multi-head attention by varying the number of heads from 2 to 8. As shown in Table~\ref{human_depth}, using 4 attention heads achieves the best gait recognition performance, with Rank-1 and Rank-10 accuracies of 68.06\% and 92.76\%, respectively. Attribute estimation results in Table~\ref{human_depth} also confirm that 4 heads achieve the most stable and accurate predictions. Fewer heads (2) reduce model expressiveness, while excessive heads (8) may lead to diluted attention and increased training difficulty.

{\bf Effect of FC Dimension in MLP (\#Dim):}
We also vary the FC dimension within the fusion layer (256, 512, 1024). As shown in Tables~\ref{human_depth}, setting the FC dimension to 512 achieves the best overall performance for both gait recognition and human attribute estimation. A smaller dimension (256) limits the model’s capacity to represent complex interactions, while a larger dimension (1024) slightly decreases accuracy, possibly due to overparameterization.

\subsection{Visualization of Complementarity between Silhouettes and 3D SMPL parameters}
To comprehensively explore the complementarity between silhouettes and 3D SMPL parameters, we perform a detailed visualization of Combo-Gait results on the BRIAR dataset, as shown in~\Cref{fig:Human Attributes Estimation Visualization}. We illustrate an example containing a probe set and a rank-1 gallery set from two distinct subjects. Five frames are sampled from each sequence. The first two rows of the visualization demonstrate the effectiveness of our approach under high-quality sample conditions. Accurate and reliable performance in these optimal scenarios underscores the robustness of Combo-Gait. The following two rows reveal that the method maintains strong performance even when parts of the subject are missing or occluded (highlighted with red boxes). This resilience suggests that the 3D SMPL parameters provide complementary information (highlighted with green boxes) that compensates for incomplete silhouette data. These results highlight the value of integrating both 2D silhouettes and 3D body parameters, confirming that the fusion of these modalities significantly enhances the robustness and overall performance of the Combo-Gait framework.

\section{Conclusion}
In this paper, we introduced Combo-Gait, a novel multi-modal and multi-task framework for gait recognition that effectively combines 2D silhouette features with 3D SMPL-based features to capture the full geometric and dynamic complexity of human walking. By integrating a multitask learning strategy for simultaneous gait recognition and human attribute estimation, our approach leverages shared representations to improve both identity discrimination and attribute prediction. The unified transformer-based fusion network enables effective integration of multi-modal features while preserving essential temporal and attribute-aware information. Extensive experiments on the challenging large-scale BRIAR dataset demonstrate that our method significantly outperforms state-of-the-art approaches, achieving superior performance in gait recognition and accurate human attribute estimation under unconstrained conditions. These results validate the effectiveness of multi-modal and multitask learning and highlight the potential of our approach to advance robust gait analysis for real-world applications such as long-range surveillance and biometric identification.

{\small
\bibliographystyle{ieee}
\bibliography{egbib}
}

\end{document}